%% file: main.tex
\documentclass[10pt,twocolumn,letterpaper]{article}

\usepackage[pagenumbers]{cvpr} 


\definecolor{cvprblue}{rgb}{0.21,0.49,0.74}
\usepackage[pagebackref,breaklinks,colorlinks,allcolors=cvprblue]{hyperref}
\usepackage{multirow}
\usepackage{pifont}
\usepackage{array}
\usepackage{adjustbox}
\newcolumntype{H}{>{\setbox0=\hbox\bgroup}c<{\egroup}@{}}

\usepackage{cuted}

\title{AerialGo: Walking-through City View Generation from Aerial Perspectives}

\author{Fuqiang Zhao$^{1, 4*}$
\and
Yijing Guo$^{1*}$
\and
Siyuan Yang$^{2*}$
\and
Xi Chen$^3$
\and
Luo Wang$^1$
\and
Lan Xu$^1$ 
\and
Yingliang Zhang$^{3}$ \qquad Yujiao Shi$^1$ \qquad Jingyi Yu$^{1}$\\
$^{1}$ ShanghaiTech University \qquad $^{2}$ Huazhong University of Science and Technology \qquad $^{3}$ DGene \qquad \\ $^{4}$ NeuDim
}
\begin{document}
\maketitle

\begin{strip}
    \centering
    \setlength{\abovecaptionskip}{1pt}
\setlength{\belowcaptionskip}{1pt}
    \includegraphics[width=1.0 \textwidth]{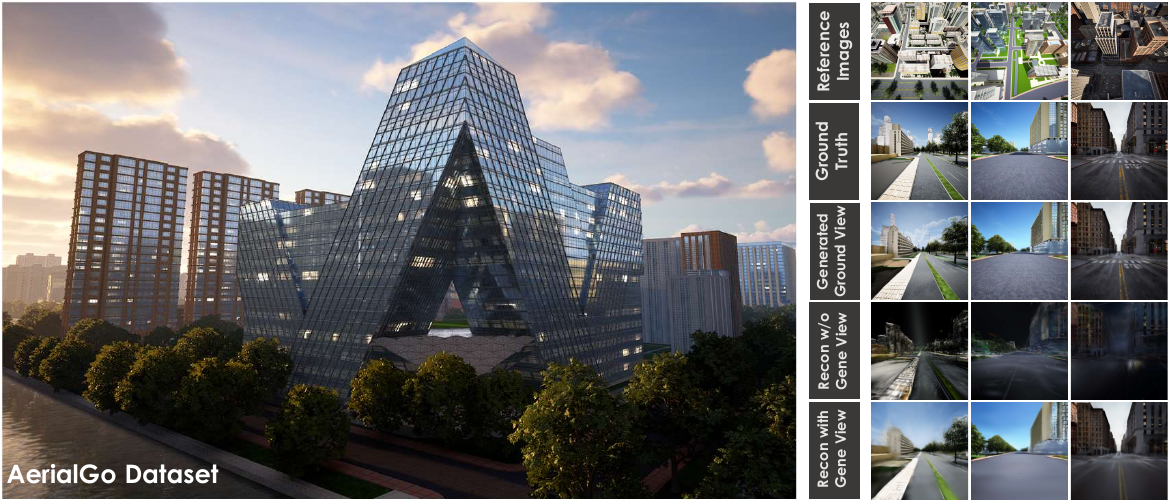}
    \captionof{figure}{\textbf{Overview of the AerialGo dataset and results.} (a) The AerialGo dataset is a large-scale, multi-view dataset, encompassing aerial and ground perspectives, and multi-attribute dataset. (b) Leveraging the AerialGo dataset, we introduce the AerialGo method, an innovative multi-view diffusion framework designed to synthesize photorealistic ground-level imagery from aerial observations, enabling enhanced urban scene reconstruction and realistic walkthrough experiences.}
    \label{fig:teaser}
    \vspace{-0.2cm}
\end{strip}
\footnote{$*$ Equal contribution}
\input{sec/0_abstract}    
\input{sec/1_intro}

\input{sec/2_related_work}

\input{sec/dataset}

\input{sec/3_method}
\input{sec/4_result}
\input{sec/5_conclusion}
{
    \small
    \bibliographystyle{ieeenat_fullname}
    \bibliography{main}
}


\end{document}

%% file: sec/0_abstract.tex
\begin{abstract}

High-quality 3D urban reconstruction is essential for applications in urban planning, navigation, and AR/VR. However, capturing detailed ground-level data across cities is both labor-intensive and raises significant privacy concerns related to sensitive information, such as vehicle plates, faces, and other personal identifiers. To address these challenges, we propose AerialGo, a novel framework that generates realistic walking-through city views from aerial images, leveraging multi-view diffusion models to achieve scalable, photorealistic urban reconstructions without direct ground-level data collection. By conditioning ground-view synthesis on accessible aerial data, AerialGo bypasses the privacy risks inherent in ground-level imagery. To support the model training, we introduce AerialGo dataset, a large-scale dataset containing diverse aerial and ground-view images, paired with camera and depth information, designed to support generative urban reconstruction. Experiments show that AerialGo significantly enhances ground-level realism and structural coherence, providing a privacy-conscious, scalable solution for city-scale 3D modeling.



\end{abstract}

%% file: sec/1_intro.tex
\section{Introduction}
\label{sec:intro}


With rapid urbanization, cities worldwide are not only functional spaces but also embody distinct cultural and architectural features, from New York’s grid layout to Beijing’s quadrangle courtyards. High-quality modeling and rendering of cities benefit numerous applications, ranging from urban planning to autonomous navigation. 
However, constructing 3D city models with high fidelity and realism poses significant challenges. Manual construction is infeasible due to the complexity and vast scale of urban environments, requiring immense time and resources to model every architectural detail.

Advances in 3D reconstruction can potentially automate the recovery of 
offer automated aps to recover 
detailed 3D structures from images. Early work by Agarwal et al.~\cite{agarwal2011building} demonstrated the feasibility of using Structure from Motion and Multi-View Stereo to ``Build Rome in a Day''. More recent approaches substitute the MVS module with differentiable alternatives such as NeRF~\cite{mildenhall2021nerf,turki2022mega} or 3D Gaussian Splatting (3DGS)~\cite{kerbl20233d} to enhance reconstruction quality and efficiency. 
Despite these achievements, they uniformly rely on
high-quality ground-level images.  Collecting such images is, however, labor-intensive, costly, and often impractical due to accessibility and regulatory constraints. ``Building Rome in a Day''~\cite{agarwal2011building} mitigates the challenge of data scarcity for landmarks by utilizing crowd-sourced images. For example, they use twenty thousand cityscape images of Rome to reconstruct the detailed 3D models. Although effective, only a few well-known cities have sufficient online ground-level image data, let alone new cities or regions under development that require comprehensive planning.
While using cars equipped with LiDar or cameras to capture images is a viable option, it is both expensive and subject to regulatory restrictions. 
In addition, ground-level images, whether sourced online or newly captured, raise significant privacy concerns: they often capture sensitive details like vehicle license plates and faces, necessitating complex anonymization.
These limitations underscore the urgent need for alternative approaches not relying on online ground-view data.

In this paper, we resort to aerial imagery to approximate ground-level scenes for urban 3D reconstruction. 
Aerial images captured by satellites, drones, and aircraft are much easier to obtain, enabling rapid and efficient coverage of vast areas without the logistical challenges of ground-level photography. Moreover, aerial imagery offers enhanced privacy protection due to higher altitude capture, making it difficult to identify personal details at ground level. 
Recent methods such as CityGaussian~\cite{liu2024citygaussian} and VastGaussian~\cite{lin2024vastgaussian} leverage aerial data with 3DGS and Level of Detail (LoD) rendering for scalable scene reconstructions. 
While these approaches excel in producing high-quality images from aerial viewpoints, they struggle to render realistic ground-level perspectives due to restricted angles and occlusions inherent in aerial imagery, reducing the overall quality and realism of pedestrian viewpoints.

To overcome the limitations of aerial-only reconstruction, we propose \textbf{AerialGo, a novel multi-view diffusion framework} that generates walking-through ground-level images from aerial imagery. 
Unlike recent works such as Reconfusion~\cite{wu2024reconfusion}, Cat3D~\cite{gao2024cat3d},  LucidDreamer~\cite{chung2023luciddreamer}, and ViewCraft~\cite{yu2024viewcrafter} which focus on object-level or limited-angle scenarios and struggle with large-scale urban environments with significant elevation differences, AerialGo effectively handles complex elevation variations, facilitating accurate and comprehensive 3D reconstructions in diverse urban settings.
AerialGo encodes aerial images and combines them with random noise for the target ground view, then processes through 3D attention to generate realistic images via denoising. To ensure 3D structural consistency for critical features like roads and buildings, we further render point cloud images based on ground camera parameters, extract features, and integrate these features as conditions for the diffusion model. This multi-faceted approach enables AerialGo to produce coherent, high-quality ground views, advancing the capabilities of urban-scale 3D reconstruction, e.g., through 3DGS, NeRF, and MVS-based approaches. 

\begin{figure*}[t]
    \centering
    \setlength{\abovecaptionskip}{0pt}
\setlength{\belowcaptionskip}{0pt}
    \includegraphics[width=1.0\linewidth]{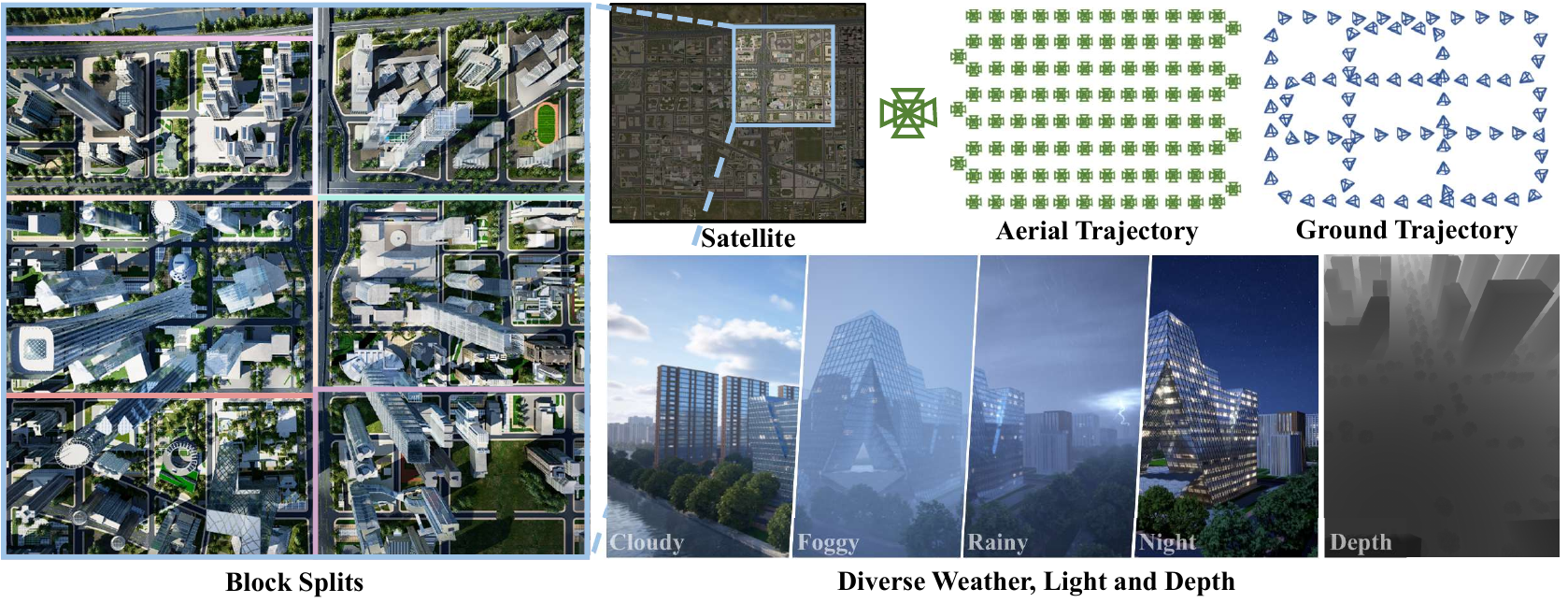}
    \caption{\textbf{Overview of the dataset and data collection process.} This figure showcases an example of our urban city model, highlighting the block partitioning, the design of aerial and ground trajectories, as well as the dynamic rendering capabilities.}
    \label{fig:dataset}
    \vspace{-0.4cm}
\end{figure*}

In addition, we present \textbf{an AerialGo dataset} that includes 3.45 million aerial and ground-view images across 134 km$^2$ of diverse urban environments, with varied lighting and weather conditions, precise camera parameters, and depth information, which will be disseminated to the research community. This dataset supports training large-scale generative models and advancing city-scale 3D reconstruction (See Table~\ref{table:dataset} and Fig.~\ref{fig:dataset}).
To our knowledge, AerialGo is the first framework to use diffusion models for city-level reconstruction by generating ground views from aerial imagery. We conduct extensive experiments across different urban datasets and compare AerialGo with various existing aerial-only and hybrid methods. 
The results demonstrate that AerialGo achieves superior ground-view fidelity and realism, especially in occluded and complex urban areas, thereby bridging the aerial-to-ground data gap for scalable, privacy-compliant 3D urban reconstruction.


Our primary contributions are as follows: 
\begin{itemize} 
\item We introduce AerialGo, a novel multi-view diffusion framework that synthesizes realistic ground-level views from aerial imagery, enabling scalable and privacy-conscious 3D city-scale reconstruction.

\item We present AerialGo dataset, a large-scale dataset with extensive aerial and ground-level images across diverse urban environments designed to support aerial-to-ground reconstruction tasks and multi-view generative modeling.

\item Through extensive experiments, we demonstrate that AerialGo achieves high-quality ground-level synthesis, significantly improving visual realism in city-scale reconstructions.
\end{itemize}

%% file: sec/2_related_work.tex
\section{Related Work}
\label{sec:related_work}

\noindent\textbf{3D Reconstruction for Large-scale Scenes.}
City-scale 3D reconstruction has a long history in photogrammetry and computer vision, originating from early techniques like Structure from Motion (SfM)~\cite{schonberger2016structure} and Multi-View Stereo (MVS)~\cite{goesele2006multi}, which rely on large-scale image datasets for 3D model building. “Building Rome in a Day”~\cite{agarwal2011building} demonstrates the feasibility of reconstructing extensive urban scenes using millions of publicly available images and SfM techniques. 
New 3D representations, such as Neural Radiance Fields (NeRF)~\cite{mildenhall2021nerf}, 3D Gaussian Splatting (3DGS)~\cite{kerbl20233d} and their extensions~\cite{chen2021mvsnerf,chen2022tensorf,muller2022instant,zhao2022human,zhao2022humannerf, cui2024letsgo,yu2024mip,barron2021mip,fridovich2022plenoxels,Reiser2021ICCV,Huang2DGS2024,zhou2024drivinggaussian,yan2024street}, have made photorealistic reconstruction more efficient and feasible. BlockNeRF~\cite{tancik2022block} introduced a divide-and-conquer strategy by segmenting city streets into blocks and applying NeRF independently to each block.
Similarly, Hierarchical 3DGS~\cite{hierarchicalgaussians24} utilizes a multi-layer structure and Level-of-Detail (LOD) rendering techniques to improve reconstruction quality.
Despite the progress, these approaches generally require extensive multi-view observations, which are challenging to acquire in urban-scale settings due to limited ground-view data. Our method alleviates this limitation by generating ground-view priors from aerial observations, enabling effective reconstruction even with sparse ground-level inputs.

 \begin{table*}[ht]
    \centering  
    \setlength{\abovecaptionskip}{2pt}
\setlength{\belowcaptionskip}{2pt}
\setlength{\tabcolsep}{6pt}
    \footnotesize
        \begin{tabular}{lHcccccccccH}
            \toprule
            Dataset & Types & Images & Area & DOF & Perspective & Aerial Viewpoint & Depth & Weather & Light & Sequence & Intrinsics \\ 
            \midrule
            UAVD4L~\cite{Wu2023uav4l} & Scene & 0.3K & \(2.5 km^2\) & 6 & UAV & Arbitrary & \ding{51} & \ding{55} & \ding{55} & \ding{51} & \ding{55} \\
            CrossLoc~\cite{yan2021crossloc} & Scene & 57K & \(2.7 km^2\) & 6 & UAV & Arbitrary & \ding{51} & \ding{55} & \ding{55} & \ding{51} & \ding{55} \\
            UrbanScene3D~\cite{UrbanScene3D} & Scene & 128K & \(55 km^2\) & 6 & UAV & Arbitrary &  \ding{51}& \ding{55} & \ding{55} & \ding{51} & \ding{55} \\
            UrbanBIS~\cite{yang2023urbanbis} & City & 113K & \(10.78 km^2\) & 3 & UAV & Arbitrary &  \ding{55}& \ding{55} & \ding{55} & \ding{51} & \ding{55} \\
            Mill 19~\cite{Turki_2022_CVPR} & Scene & 3.5K & \(0.18 km^2\) & 6 & UAV & Overhead &  \ding{55}& \ding{55} & \ding{51} & \ding{51} & \ding{51} \\
            UAV-VisLoc~\cite{xu2024uavvisloc} & Scene & 7K & / & 6 & Satellite+UAV & Overhead &  \ding{55}& \ding{55} & \ding{55} & \ding{51} & \ding{55} \\
            Dense UAV~\cite{DenseUAV} & City & 27K & / & 3 & Satellite+UAV & Overhead &  \ding{55}& \ding{51} & \ding{51} & \ding{51} & \ding{55} \\
            \midrule
            CVUSA~\cite{zhai2017predicting} & City & 88K & / & 3 & Satellite+Ground & Overhead &  \ding{55}& \ding{55} & \ding{55} & \ding{55} & \ding{55} \\
            CVACT~\cite{liu2019lending} & City & 256K & / & 3 & Satellite+Ground & Overhead &  \ding{55}& \ding{55} & \ding{55} & \ding{55} & \ding{55} \\
            VIGOR~\cite{zhu2021vigor} & City & 329K & / & 3 & Satellite+Ground & Overhead &  \ding{55}& \ding{55} & \ding{55} & \ding{55} & \ding{55} \\
            GeoContext-v1~\cite{xu2024geospecific} & City & 26K & / & 3 & Satellite+Ground & Overhead &  \ding{51}& \ding{55} & \ding{55} & \ding{55} & \ding{55} \\
            \midrule
            MatrixCity~\cite{li2023matrixcity} & City & 519K & \(28 km^2\) & 6 & UAV+Ground & Arbitrary &  \ding{51}& \ding{51} & \ding{51} & \ding{51} & \ding{51} \\
            UC-GS~\cite{zhang2024drone} & City & 7K & / & 6 & UAV+Ground & Overhead & \ding{55}& \ding{55} & \ding{55} & \ding{51} & \ding{51} \\
            \midrule
            \textbf{AerialGo} & City & \textbf{3.45M} & \textbf{134\(km^2\)} & 6 & UAV+Ground & Arbitrary &  \ding{51}& \ding{51} & \ding{51} & \ding{51} & \ding{51} \\
            \bottomrule
        \end{tabular}

    \rule{0pt}{0.01pt}
    \caption{Comparison of the statistics and key attributes of our dataset against existing datasets.}
    \label{table:dataset}
    \vspace{-0.6cm}
\end{table*}

\noindent \textbf{Exploring Diffusion Priors for 3D Reconstruction.}
%
DreamFusion~\cite{poole2022dreamfusion} utilizes Score Distillation Sampling (SDS) to reconstruct NeRF solely based on a text prompt, while other methods~\cite{shi2023mvdream,wang2023imagedream,shi2023zero123++,liu2023syncdreamer} fine-tune text-to-image latent diffusion models to generate coherent multi-view images simultaneously. For instance, ImageDream~\cite{wang2023imagedream} captures multi-view dependencies using a diffusion architecture similar to video diffusion models, employing 3D self-attention mechanisms. 
Video diffusion models~\cite{ho2022video,ho2022imagen,blattmann2023stable,jain2024video,wang2024motionctrl} have shown remarkable capabilities in generating consistent videos, and are believed to implicitly reason about 3D structures. 
For example, MotionCtrl~\cite{wang2024motionctrl} conditions the diffusion model on arbitrarily specified camera trajectories, while ViewCraft~\cite{yu2024viewcrafter} additionally conditions on images rendered from a coarse 3D point cloud to generate high-quality video frames. However, these methods are typically constrained to small view ranges and struggle to handle the significant viewpoint transitions. 
In contrast, our AerialGo method explicitly models the large viewpoint disparity between aerial and ground views, leveraging point cloud information and camera embeddings to provide strong 3D structure conditioning. This allows our model to effectively bridge the gap between aerial observations and ground-level reconstructions.

\noindent \textbf{Satellite-to-Ground View Generation.}
Cross-view synthesis, particularly satellite-to-ground view generation, aims to produce ground-level images from a significantly different aerial or top-view satellite image. 
%
%
Several works~\cite{zhai2017predicting,toker2021coming,wu2022cross,lu2020geometry, shi2022geometry,qian2023sat2density,li2021sat2vid, li2024sat2scene,xu2024geospecific} have addressed these challenges by leveraging geometric priors, multi-view consistency, and advanced generative models. The pioneering work in this domain, S2G~\cite{lu2020geometry}, introduced geometry consistency by predicting a density voxel grid from satellite height maps, transforming it into ground-level depth and semantic panorama, followed by a 2D generative model to create the ground view. Building on this, Sat2Vid~\cite{li2021sat2vid} addressed the challenge of temporal consistency when extending satellite-to-ground synthesis to video generation. 
More recent work, InfiniCity~\cite{lin2023infinicity}, scaled satellite-to-ground view generation to a city-wide level. 
On the other hand, Sat2Density~\cite{qian2023sat2density} focuses on predicting top-view density maps without explicit depth supervision, leveraging the relationship between satellite and ground views. It integrates neural rendering techniques to enhance the visual quality of synthesized views but still struggles with maintaining temporal consistency in generated video sequences.

%% file: sec/dataset.tex
\section{AerialGo Dataset}

We review several existing aerial datasets with posed aerial images that could potentially support city or 3D reconstruction, detailed in Table~\ref{table:dataset}. Datasets such as UAVD4L~\cite{Wu2023uav4l}, CrossLoc~\cite{yan2021crossloc}, UrbanScene3D~\cite{UrbanScene3D}, UrbanBIS~\cite{yang2023urbanbis}, Mill 19~\cite{Turki_2022_CVPR}, UAV-VisLoc~\cite{xu2024uavvisloc}, and DenseUAV~\cite{DenseUAV} contain only aerial images without ground-view data, limiting their applicability in comprehensive city-level modeling. Other datasets, like CVUSA~\cite{zhai2017predicting}, CVACT~\cite{liu2019lending}, VIGOR~\cite{zhu2021vigor}, and GeoContext-v1~\cite{xu2024geospecific}, focus on satellite-to-ground view generation but lack comprehensive 3D reconstruction labels. While MatrixCity~\cite{li2023matrixcity} and UC-GS~\cite{zhang2024drone} are closer to our goals, their relatively limited scale and architectural diversity make them difficult to train a diffusion model effectively. 
Our AerialGo dataset addresses these limitations by including approximately 3.45 million images with coverage of 134 km$^2$ from different urban models in Unreal Engine 5 (UE5), encompassing both aerial and ground views. 
%
AerialGo also provides camera parameters, depth maps, and segmentation maps for each view. Furthermore, we offer flexible control over images under various environmental conditions, including different weather, lighting directions, and times of day, enabling simulations of dynamic real-world scenarios. 

\begin{figure*}[t]
    \centering
    \setlength{\abovecaptionskip}{0pt}
\setlength{\belowcaptionskip}{0pt}
    \includegraphics[width=1.0\linewidth]{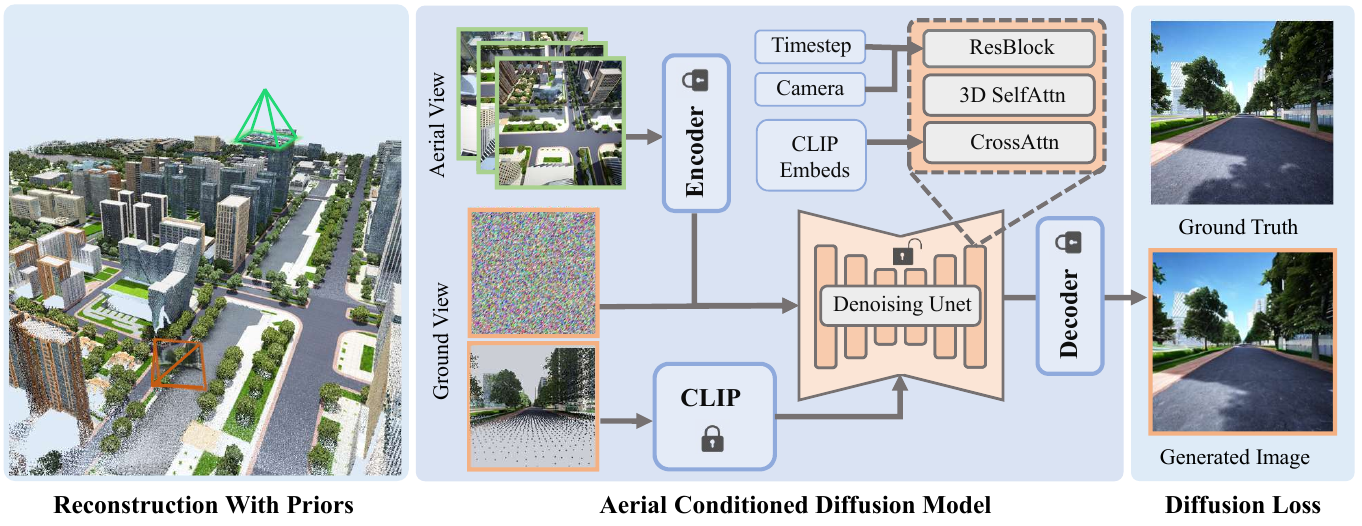}
    \caption{\textbf{Pipeline of the AerialGo method.} Starting with a target ground view, we first select reference images from the nearest aerial views and encode them using a pretrained auto-encoder. The diffusion model then processes the encoded aerial features along with random noise at the ground view, passing the adapted features through 3D self-attention layers. Additionally, CLIP embeddings of the ground-view point cloud render are integrated via cross-attention layers to enhance structural consistency in the generated views. The resulting priors contribute to improved 3D urban reconstruction quality, especially at ground level.}
    \label{fig:pipeline}
    \vspace{-0.4cm}
\end{figure*}

\subsection{Scale and Diversity}
\textbf{Area.}
Our dataset spans different urban areas, covering a total of 134 km². These regions feature diverse terrains, building heights, and architectural styles, providing rich training data for large-scale 3D urban reconstruction, rendering, and scene generation. The variety in layouts includes high-rise buildings, residential neighborhoods, commercial zones, and complex urban scenes such as plazas, parks, and transit hubs, making the dataset suitable for a wide range of tasks across various environments. Notably, many of the buildings are characterized by reflection glass curtainwalls, which introduce complex lighting reflections influenced by different weather conditions and times of day. This adds an additional layer of realism and variability to the dataset, further enhancing its value for training models in dynamic urban settings.

\noindent\textbf{Dynamic Environments.}
To simulate realistic urban dynamics, we collect data under various weather conditions (e.g., sunny, rainy, foggy) and lighting scenarios (e.g., daytime, sunset, nighttime). Additionally, we control time of day, light angle, and intensity to capture natural lighting shifts in a day. This control enables the generation of diverse, dynamic scenes that align with real-world application needs, supporting more lifelike city-scale neural renderings.

\subsection{Data Collection}
\textbf{Aerial view collection.}
To efficiently capture aerial data of the city, we employ an oblique photography approach. A custom-developed trajectory generation script automates the creation of camera paths for each block, eliminating the need for manual annotation and greatly enhancing collection efficiency. Fig.~\ref{fig:dataset} shows the fully continuous aerial and ground trajectories for a city block, enabling complete traversability unlike existing piecewise-continuous datasets. We adjust the Unmanned Aerial Vehicle (UAV)'s flight altitude and trajectory sampling density according to building height and density. The UAV is equipped with five synchronized cameras: one downward-facing camera and four oblique cameras angled 60 degrees downwards from the horizon, facing forward, backward, left, and right relative to the UAV. This setup ensures comprehensive coverage of building and street details. Additionally, we simulate realistic effects such as ray tracing, motion blur, and anti-aliasing to enhance the realism of the rendered images, producing high-quality city-scale data suitable for neural rendering tasks.

\textbf{Ground view collection.}
For ground-view data collection, we manually annotate the start and end points of each road, using these as guides for automatically generated straight-line paths. To ensure smooth turns at intersections and avoid collisions with buildings, we add additional key points at turning locations, enabling safe and smooth trajectories. The generated camera trajectories are saved, allowing for consistent re-rendering under different environmental conditions. 
Ground-view and aerial-view images inherently require different exposure settings due to varying lighting conditions and perspectives. Therefore, we apply automatic exposure adjustments specific to each view type. This approach ensures realistic lighting representation across both ground and aerial views, capturing the distinct visual characteristics of each perspective rather than enforcing a uniform exposure.



%% file: sec/3_method.tex
\section{The Proposed AerialGo Framework}
\label{method}

\textbf{AerialGo} is designed to generate walking-through city views from aerial images and facilitate novel view synthesis from both aerial and ground-level perspectives. The overall pipeline is illustrated in Fig.~\ref{fig:pipeline}. Our methodology comprises two main components: the diffusion pipeline (Sec.~\ref{diffusion}) and the validation of large-scale Gaussian reconstruction methods using the generated walk-through view priors (Sec.~\ref{3dgs}).

\subsection{Diffusion Model for Aerial2Ground Generation}
\label{diffusion}

Urban scene reconstruction often suffers from the scarcity of ground-level data, which limits the quality of ground-view renderings. To address this, we introduce a novel diffusion model that generates ground-view images conditioned on aerial images and reconstructed 3D point clouds from them. 
The substantial angular disparity between aerial and ground views poses significant challenges for existing novel view synthesis (NVS) methods by leveraging diffusion priors, which typically focus on objects or small-scale scenes with limited view ranges, such as ImageDream~\cite{wang2023imagedream}, LucidDreamer~\cite{chung2023luciddreamer}, ReconFusion~\cite{yu2024viewcrafter}, and ViewCraft~\cite{wu2024reconfusion}.

Given aerial image data \( I_A = \{I_i\}_{i=0}^K \), we first estimate the camera parameters \( C_A = \{C_i\}_{i=0}^K \) and reconstruct the corresponding 3D point cloud \( P \). For a target ground-view camera \( C_G \), we select the \( N \) closest reference aerial images \( I_R = \{I_i\}_{i=0}^N \subseteq I_A \) and their camera parameters \( C_R = \{C_i\}_{i=0}^N \) based on the position and orientation of \( C_G \). The 3D point cloud is then rendered from the ground-view camera parameters to produce the point cloud rendering \( I_P \), which serves as a 3D structural condition. The diffusion model aims to learn the distribution of ground-view images conditioned on aerial views, expressed as \( p(I_G | I_R, C_R, C_G, I_P ) \).

\textbf{Selection of Reference Aerial Images.} In an oblique photogrammetry setup, aerial drones typically employ five cameras oriented front, back, left, right, and downward. Given a ground camera parameter \( C_G \), we identify the drone closest to the ground camera's location. Based on \( C_G \)'s orientation, we select the \( N = 3 \) aerial images with the most similar directional angles, ensuring the inclusion of the downward-facing image for comprehensive coverage.

\textbf{Diffusion Model Architecture.} Our diffusion model builds upon existing multi-view and video diffusion frameworks (see Fig.~\ref{fig:pipeline}), with a key modification: each denoising step focuses exclusively on generating ground-view images. Specifically, given \( N \) reference aerial-view images of resolution \(256 \times 256 \times 3\), these images are processed using a pre-trained autoencoder to extract latent representations of shape \( N \times 32 \times 32 \times 4 \).
During training, the aerial-view latents are concatenated with the latent representation of the target ground-view image after adding noise, resulting in a combined input of shape \( (N+1) \times 32 \times 32 \times 4 \). This input is fed into the diffusion model. During inference, instead of the ground-view latent, randomly sampled noise is concatenated with the aerial latents to generate the ground-view image.
Our model employs a 3D self-attention mechanism, where 1D attention captures information across multiple images and 2D attention captures intra-image features. This design effectively leverages multi-view aerial context to synthesize consistent ground-level imagery.

Our model is initialized with parameters from ImageDream~\cite{wang2023imagedream}, pre-trained on the Objaverse~\cite{deitke2023objaverse} dataset. While Objaverse focuses on foreground objects, our urban scenes require a balanced representation of complex backgrounds and foreground elements. To accommodate this, we increase the noise scale at each timestep during DDIM sampling, allowing the model to explore a broader latent space and enhancing generalization while mitigating overfitting to object-centric features.

To ensure content consistency between aerial and ground views, we incorporate camera parameters and 3D point clouds reconstructed from aerial images as 3D structural conditions in the diffusion model. Specifically, the point cloud rendering of the ground view is encoded using a pre-trained CLIP encoder to obtain CLIP embeddings. The camera parameters are projected to match the dimension of the time embeddings to obtain camera embeddings. These embeddings are combined and fed into the residual network of the diffusion model. Additionally, the CLIP embeddings of the point cloud render are integrated through a cross-attention mechanism to enhance the 3D structural consistency of the generated images.

\begin{figure*}[t]
    \centering
    \setlength{\abovecaptionskip}{1pt}
\setlength{\belowcaptionskip}{1pt}
    \includegraphics[width=0.95\linewidth]{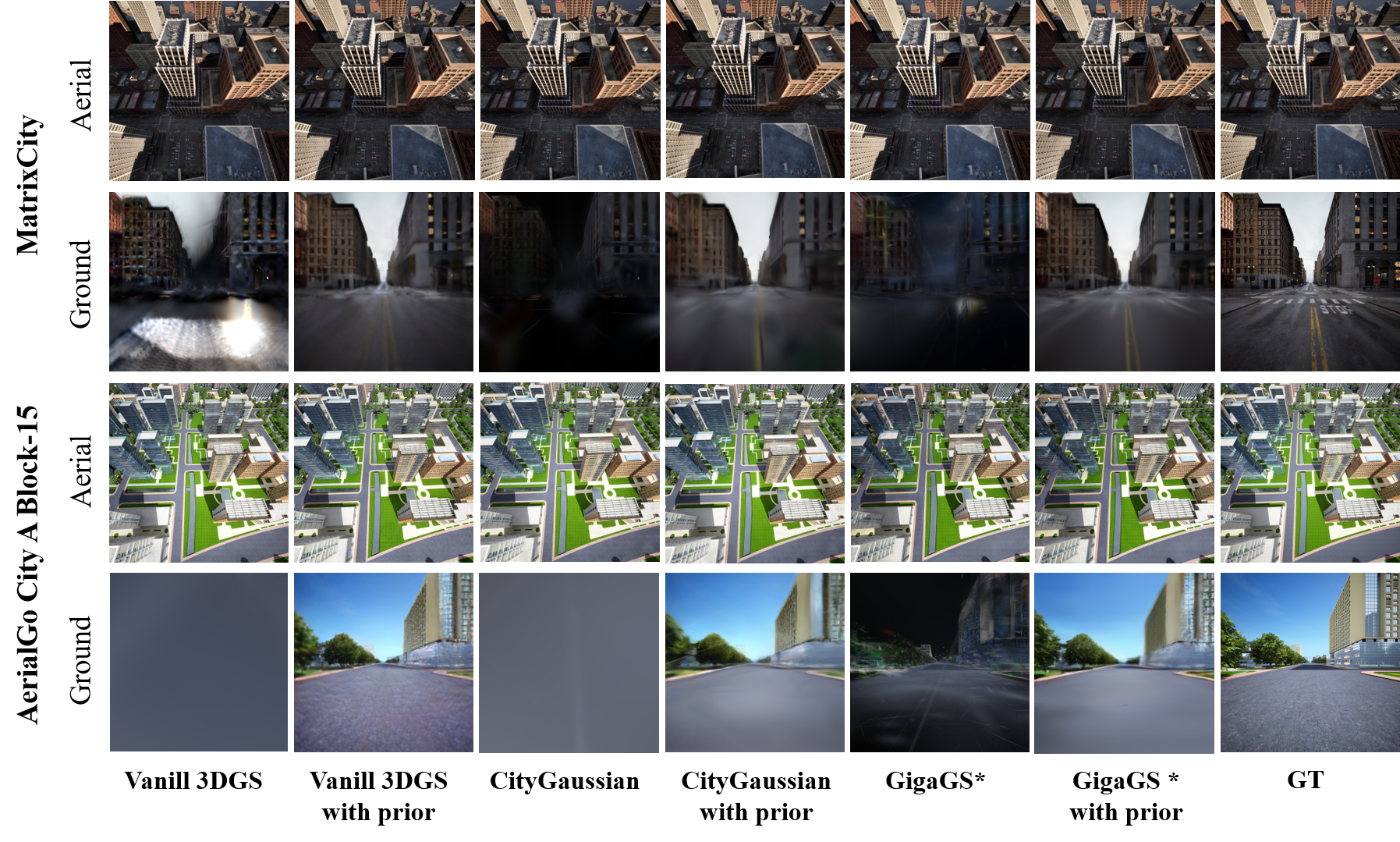}

    \caption{\textbf{Qualitative comparison of 3D reconstruction methods with or without our generated ground view priors.} * notes that the method is implemented by ourselves.}
    \label{fig:comparison}
    \vspace{-0.3cm}
\end{figure*}

\subsection{3D Reconstruction with Diffusion Priors}
\label{3dgs}

State-of-the-art 3D reconstruction methods, such as NeRF and 3DGS, typically require abundant multi-view observations to achieve high-quality results. However, acquiring extensive ground-level data for urban scenes is labor-intensive and challenging. Our proposed diffusion model addresses this limitation by generating ground-view images from aerial images, thereby reducing the dependency on manual ground-level data collection.

The aerial-view generation model produces high-quality ground views that demonstrate great consistency in low-frequency regions, such as road surfaces. Nonetheless, in the high-frequency areas with complex textures, including building facades and tree leaves, discrepancies in appearance details persist, even though the 3D structural integrity is maintained. This highlights a persistent challenge for current diffusion models in accurately rendering detailed textures in complex urban environments.

To effectively utilize and validate the generated ground-view priors, we integrate them into multiple 3DGS backbone methods. Each method is enhanced by incorporating a perceptual loss, specifically the Learned Perceptual Image Patch Similarity (LPIPS), in addition to the standard photometric reconstruction loss. Unlike photometric loss, LPIPS focuses on high-level semantic similarity between rendered and observed images, thereby mitigating inconsistencies in low-level high-frequency details.

Furthermore, we introduce skybox modeling into each reconstruction method to improve ground-view rendering performance. Specifically, we define a skybox that surrounds the extent of the scene, i.e.,
 $100k$ 3DGS primitives on a sphere $10\times $ the diameter of the scene, to capture the effect of the sky. And we only optimize opacity and SH coefficients of 3D Gaussians of skybox. This addition enhances the overall stability and realism of ground-level perspectives by providing consistent environmental context.

This comprehensive approach leverages the strengths of our diffusion-generated priors while addressing their limitations, resulting in more accurate and visually coherent 3D urban reconstructions.

\begin{table*}[t]
	\centering
    \setlength{\abovecaptionskip}{2pt}
\setlength{\belowcaptionskip}{2pt}
\small
	\begin{tabular}{lccccccccccc}
		\toprule
		\multirow{2}{*}{Method} & \multirow{2}{*}{Settings} & \multicolumn{3}{c}{AerialGo City A}&\multicolumn{3}{c}{AerialGo City B} & \multicolumn{3}{c}{MatrixCity Dataset} \\
		\cmidrule(lr){3-5}\cmidrule(lr){6-8} \cmidrule(lr){9-11}
		
		&& \multicolumn{1}{c}{PSNR$\uparrow$} & \multicolumn{1}{c}{SSIM$\uparrow$} & \multicolumn{1}{c}{LPIPS$\downarrow$} &
		\multicolumn{1}{c}{PSNR$\uparrow$} & 
		\multicolumn{1}{c}{SSIM$\uparrow$} & 
		\multicolumn{1}{c}{LPIPS$\downarrow$} &\multicolumn{1}{c}{PSNR$\uparrow$} & 
		\multicolumn{1}{c}{SSIM$\uparrow$} & 
		\multicolumn{1}{c}{LPIPS$\downarrow$}\\
		\midrule
		 3DGS & \multirow{3}{*}{\shortstack{Recon \\ w/o priors}}  & 19.69 & 0.686  & 0.328   & 16.13 & 0.589 & 0.398  & 19.16 & 0.676 & 0.404    \\
		 CityGaussian  &  &19.25 & 0.674 & 0.350 &15.92 & 0.587  &0.404  & {21.84} & {0.671} & {0.333}     \\
		GigaGS* &  & 17.36 & 0.627 & 0.330 & 15.11 & 0.532  & \textbf{0.351} &  {21.56} & {0.711} & {0.312}  \\
            \midrule
            3DGS & \multirow{03}{*}{\shortstack{Recon\\ with priors}}  & 20.13 & \textbf{0.626} & 0.397 & \textbf{17.28}& \textbf{0.625}  & 0.379  &  23.82 & 0.756 & 0.376 \\
		 CityGaussian  &  & \textbf{20.27} & 0.654 & 0.387 & 16.78 & 0.578  &0.448 &  
         24.75 & 0.788 & 0.345
         \\
		GigaGS*   &   & 20.21 & 0.682 & \textbf{0.320} & 16.71 & 0.622  & 0.372  & \textbf{25.52}& \textbf{0.800} & \textbf{0.311} \\
		\bottomrule
	\end{tabular}
\rule{0pt}{0.01pt}
\caption{\textbf{Quantitative comparison of methods with and without our generated ground view priors.} We evaluate 3DGS~\cite{kerbl20233d}, CityGaussian~\cite{huang2022city3d}, and GigaGS~\cite{chen2024gigags} baselines on AerialGo and MatrixCity~\cite{li2023matrixcity} datasets, averaging metrics on aerial and ground views.}
\label{table:quantitative comparison}
\vspace{-0.4cm}
\end{table*}

\begin{figure*}[t]
    \centering
    \setlength{\abovecaptionskip}{0pt}
\setlength{\belowcaptionskip}{0pt}
    \includegraphics[width=1.0\linewidth]{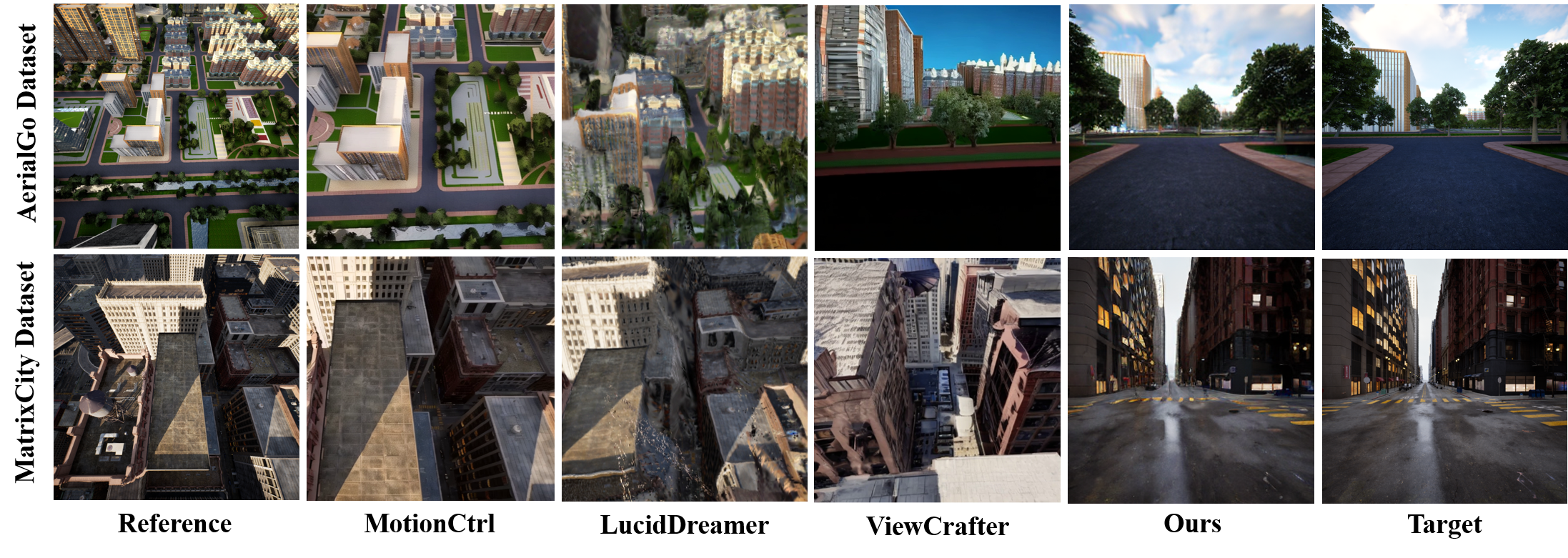}
    \caption{\textbf{Qualitative comparison of generative NVS methods on AerialGo and MatrixCity Dataset}. Compared with MotionCtrl~\cite{wang2024motionctrl}, LucidDreamer~\cite{chung2023luciddreamer}, and ViewCraft~\cite{yu2024viewcrafter}, our results aligns well with the target image.}
    \label{fig:comparison nvs}
    \vspace{-0.4cm}
\end{figure*}

\subsection{Implementation Details}
\label{implementation}

Our diffusion model is based on \textit{ImageDream} and retains its pre-trained parameters from multi-view datasets. The input images have a resolution of \(256 \times 256 \times 3\) and are downsampled to \(32 \times 32 \times 4\) using a pre-trained autoencoder before being processed by the denoising UNet.
Each training batch comprises four frames: three encoded aerial-view latents and one ground-view latent. During training, the ground-view latent is a noise-added version of the target image, whereas during inference, it is replaced with random noise. Unlike traditional diffusion models that utilize text prompts for conditioning, our model employs the CLIP embeddings of the ground-view point cloud renderings as the conditioning input. To implement classifier-free guidance (CFG), we randomly set the conditioning input to zero with a probability of 10\% and apply CFG with a scale factor of 5.0.
The diffusion model was trained on 8 NVIDIA A100 GPUs with a batch size of 64. The initial learning rate was set to \(1 \times 10^{-5}\) and was decreased to \(2.5 \times 10^{-6}\) every 3,000 iterations before being reset back to \(1 \times 10^{-5}\). The training process required approximately two days to complete.

\begin{figure}[t]
    \centering
    \setlength{\abovecaptionskip}{0pt}
\setlength{\belowcaptionskip}{0pt}
    \includegraphics[width=1.0\linewidth]{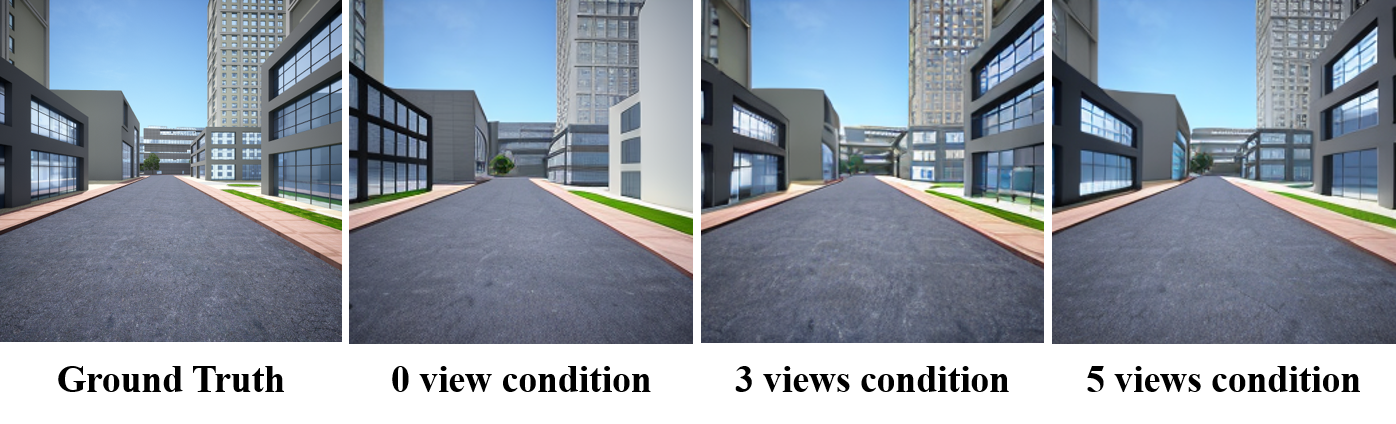}
    \caption{\textbf{Qualitative ablation on different numbers of condition views.} We found that conditioning on three aerial views provides the best result.}
    \label{fig:ablation view}
    \vspace{-0.5cm}
\end{figure}

\begin{table}[t]
	\centering
        \setlength{\abovecaptionskip}{2pt}
\setlength{\belowcaptionskip}{2pt}
\footnotesize
	\begin{tabular}{lccc}
		\toprule
		\multirow{1}{*}{Method} & \multicolumn{3}{c}{AerialGo/MatrixCity Dataset}\\
		\cmidrule(lr){2-4}
		& \multicolumn{1}{c}{PSNR$\uparrow$} & \multicolumn{1}{c}{SSIM$\uparrow$} & \multicolumn{1}{c}{LPIPS$\downarrow$}\\
		\midrule
		 LucidDreamer &  8.68/10.46 & 0.187/0.199  & 0.751/0.645   \\
		 ViewCraft  &  9.79/8.54 & 0.174/0.151 & 0.654/0.662   \\
	   MotionCrtl &  7.98/10.43 & 0.192/0.217 & 0.739/0.642    \\
          Ours &  \textbf{15.74/23.88} &  \textbf{0.491/0.722} &  \textbf{0.344/0.167}   \\
		\bottomrule
	\end{tabular}
\rule{0pt}{0.01pt}
\caption{\textbf{Comparison with other generative NVS approaches.} 
Our method shows the best performance.
}
\label{table:diffusion comparison}
\vspace{-0.6cm}
\end{table}

 \begin{table}[ht]
	\centering
        \setlength{\abovecaptionskip}{2pt}
\setlength{\belowcaptionskip}{2pt}
\footnotesize
	\begin{tabular}{lcccc}
		\toprule
		Setup & PSNR$\uparrow$ &SSIM$\uparrow$& LPIPS$\downarrow$ \\
		\midrule
            w/o point render condition  &  13.47 & 0.385 & 0.361 \\
            0-view condition & 15.71  & 0.427 & 0.313\\
		  5-views condition &  17.48  &  0.480 & 0.255 \\
            Ours (3-views condition)  &  \textbf{20.73}  &  \textbf{0.587} & \textbf{0.192}\\
		\bottomrule
	\end{tabular}
\rule{0pt}{0.01pt}
\caption{\textbf{Ablation study results.} 
We ablate two aspects of our diffusion model: the number of conditioning views and the point rendering condition. 
}
\label{table:quantitative ablation}
\vspace{-0.6cm}
\end{table}

%% file: sec/4_result.tex
\section{Experiments} 
\label{results} 

We train our diffusion model on approximately 90\% of the city blocks from the AerialGo and MatrixCity~\cite{li2023matrixcity} datasets, reserving the remaining 10\% for evaluation. The experiments demonstrate the effectiveness of AerialGo in urban 3D reconstruction by leveraging the generated ground-view images (see Sec.\ref{comparison}). Additionally, we conduct ablation studies to assess the contribution of key components in our diffusion model (see Sec.\ref{ablation}).

\subsection{Comparisons}
\label{comparison} 
We first evaluate large-scale city reconstruction using both aerial-only images and aerial-ground image pairs, where ground-view images are generated by our diffusion model. Our experiments involve three representative methods: 3DGS~\cite{kerbl20233d}, CityGaussian~\cite{huang2022city3d}, and GigaGs~\cite{chen2024gigags}, with and without the integration of our diffusion priors.

The quantitative results on the AerialGo and MatrixCity datasets are presented in Table~\ref{table:quantitative comparison}, and qualitative visualizations are shown in Fig.~\ref{fig:comparison}. When relying solely on aerial-view images, all three methods generate realistic aerial-level perspectives but struggle to render accurate ground-level views. In contrast, by incorporating ground-view images generated by our diffusion model, these methods achieve significantly improved reconstructions from the ground-level perspective.

To further demonstrate the effectiveness of our diffusion model pipeline for aerial-to-ground image generation, we compare our approach with existing methods that leverage diffusion models for novel view synthesis, including MotionCtrl~\cite{wang2024motionctrl}, LucidDreamer~\cite{chung2023luciddreamer}, and ViewCraft~\cite{yu2024viewcrafter}. For a fair evaluation in our challenging aerial-to-ground setting, we implement the necessary adaptations to these methods. As shown in Table~\ref{table:diffusion comparison} and Fig.\ref{fig:comparison nvs}, our generated results align closely with the target ground-view images, whereas other methods struggle with significant viewpoint disparity. These methods, originally designed for small scenes or limited view ranges, face difficulties in generating accurate content for large-scale urban environments.

\subsection{Ablation Studies}
\label{ablation}

\begin{figure}[t]
    \centering
    \setlength{\abovecaptionskip}{0pt}
\setlength{\belowcaptionskip}{0pt}
    \includegraphics[width=0.8\linewidth]{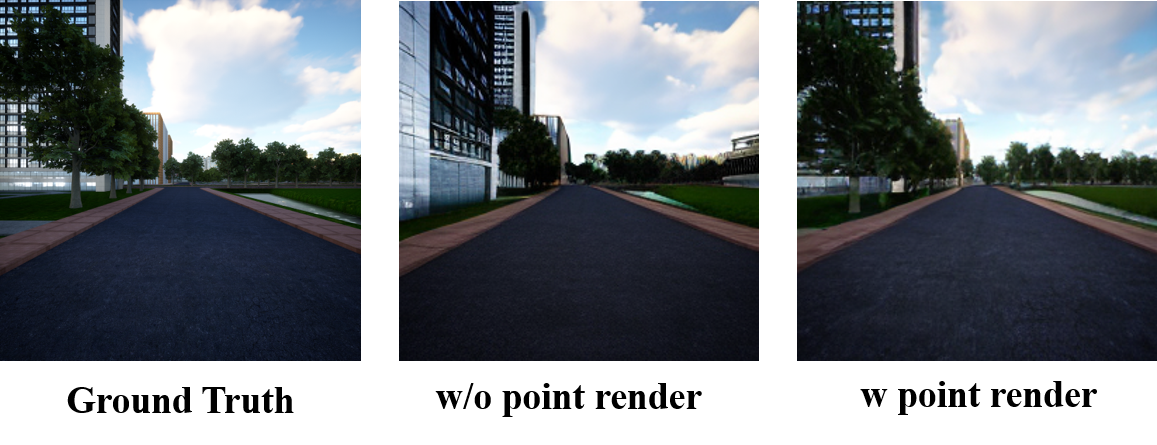}
    \caption{\textbf{Qualitative ablation on the point render condition.} Incorporating the CLIP embedding of the point rendering improves consistency with the ground truth.}
    \label{fig:ablation point}
    \vspace{-0.5cm}
\end{figure}

In this section, we conduct ablation studies on two key components of our AerialGo model: the number of conditioning views and the use of point cloud rendering embeddings. 
The results are presented in Table~\ref{table:quantitative ablation}, Fig.~\ref{fig:ablation view}, and Fig.~\ref{fig:ablation point}. 

To analyze the impact of conditioning views, we train our diffusion model with varying numbers of input aerial views, all selected from the nearest aerial drone captures. The results indicate that the model without conditioning views can still capture reasonable scene structure but struggles to produce consistent appearances compared to ground truth. On the other hand, conditioning on five aerial views, some of which may not be relevant to the target ground view, leads to inconsistent details in close-up renderings. We observe that using three aerial views as conditions strikes the best balance, providing the most consistent results both visually and quantitatively. 

Additionally, we investigate the effect of incorporating point cloud rendering conditions by removing the CLIP embedding of the point cloud rendering from the model. The results highlight the importance of the CLIP embedding, showing that its inclusion significantly improves consistency and alignment with the ground truth. 


%% file: sec/5_conclusion.tex
\section{Conclusion}
\label{conclusion}
This paper has presented AerialGo, a novel multi-view diffusion framework designed to bridge the gap between aerial imagery and ground-level 3D urban reconstruction. By leveraging aerial views to generate high-quality ground-level images, AerialGo addresses critical challenges in large-scale city modeling, including the limited availability of ground-level data and privacy concerns. Our AerialGo includes a distinctive 3D attention mechanism and point cloud-based conditioning to ensure structural consistency, producing realistic and coherent ground-level imagery. Additionally, we introduce the AerialGo dataset, a diverse and comprehensive collection of aerial and ground-view images with detailed annotations, which provides valuable resources for training generative models and advancing city-scale 3D reconstruction. Extensive experiments demonstrate that AerialGo significantly outperforms existing methods, delivering superior fidelity and realism in complex urban environments. 